\def\year{2021}\relax
\documentclass[letterpaper]{article} 
\usepackage{aaai21}  
\usepackage{times}  
\usepackage{helvet} 
\usepackage{courier}  
\usepackage[hyphens]{url}  
\usepackage{graphicx} 
\usepackage{natbib}  
\usepackage{caption} 
\usepackage[switch]{lineno} 
\usepackage{shortcuts}
\usepackage{booktabs}
\usepackage[ruled,vlined]{algorithm2e}

\title{Bayesian Variational Optimization for Combinatorial Spaces}

\iftrue
\author {
    Tony C Wu\footnote{Equal contributions.},
    Daniel Flam-Shepherd\footnotemark[1],
    Al\'{a}n Aspuru-Guzik \\
}
\affiliations {
    University of Toronto \\
    Vector Institute \\
    tonyc.wu@utoronto.ca, danielfs@cs.toronto.edu, alan@aspuru.com
}
\fi

\begin{document}
\copyrightyear{2021}
\nocopyright
\maketitle

\begin{abstract}
This paper focuses on Bayesian Optimization in combinatorial spaces. In many applications in the natural science. Broad applications include the study of molecules, proteins, DNA, device structures and quantum circuit designs, a on optimization over combinatorial categorical spaces is needed to find optimal or pareto-optimal solutions. However, only a limited amount of methods have been proposed to tackle this problem. Many of them depend on employing Gaussian Process for combinatorial Bayesian Optimizations. Gaussian Processes suffer from scalability issues for large data sizes as their scaling is cubic with respect to the number of data points. This is often impractical for optimizing large search spaces. Here, we introduce a variational Bayesian optimization method that combines variational optimization and continuous relaxations to the optimization of the acquisition function for Bayesian optimization. Critically, this method allows for gradient-based optimization and has the capability of optimizing problems with large data size and data dimensions. We have shown the performance of our method is comparable to state-of-the-art methods while maintaining its scalability advantages. We also applied our method in molecular optimization.
\end{abstract}

\section{Introduction}

Bayesian optimization (BO) is a powerful framework for tackling global optimization problems 
involving black-box functions \cite{Jones_1998}.  BO seeks to identify an optimal solution with the minimal possible incurred costs. 
It has been widely applied, yielding impressive results on many problems in different areas ranging from
automatic chemical design \cite{G_mez_Bombarelli_2018} to hyperparameter optimization \cite{snoek2012practical} have been reported.

However, this is not true for all types of search spaces, in particular discrete spaces. Consider for example optimizing some black box function 
on a discrete grid of integers like in Figure 1. 
In this work we focus on Bayesian optimization of objective functions on combinatorial search spaces consisting of 
discrete variables where the number of possible configurations quickly explodes.
For $n$ categorical variables with $k$ categories the number of possible combinations scales with $\mathcal O(k^n)$.

Combinatorial BO \cite{baptista2018bayesian} aims to find the global optima of highly non-linear, 
black-box objectives for which simple and exact solutions are inaccurate and gradient-based optimizers are not amenable. 
These objectives typically have expensive and noisy evaluations and thus require optimizers with high sample efficiency. 
Some simple common examples of typical combinatorial optimization problems include the traveling salesman problem, integer linear programming, boolean satisfiability and scheduling.

The vast majority of the BO literature focuses on continuous search spaces. The reason for this is that BO 
relies on Gaussian processes and the smoothness from kernel methods used to model functional uncertainty. 
One first specifies a "belief" over possible explanations of the underlying function $f$
using a probabilistic surrogate model and then combines this with the use of an acquisition function 
which assesses the expected utility of a set of novel $\B X$ chosen by solving an inner optimization problem.  

\begin{figure}[t]
\centering
\includegraphics[width=0.9\columnwidth]{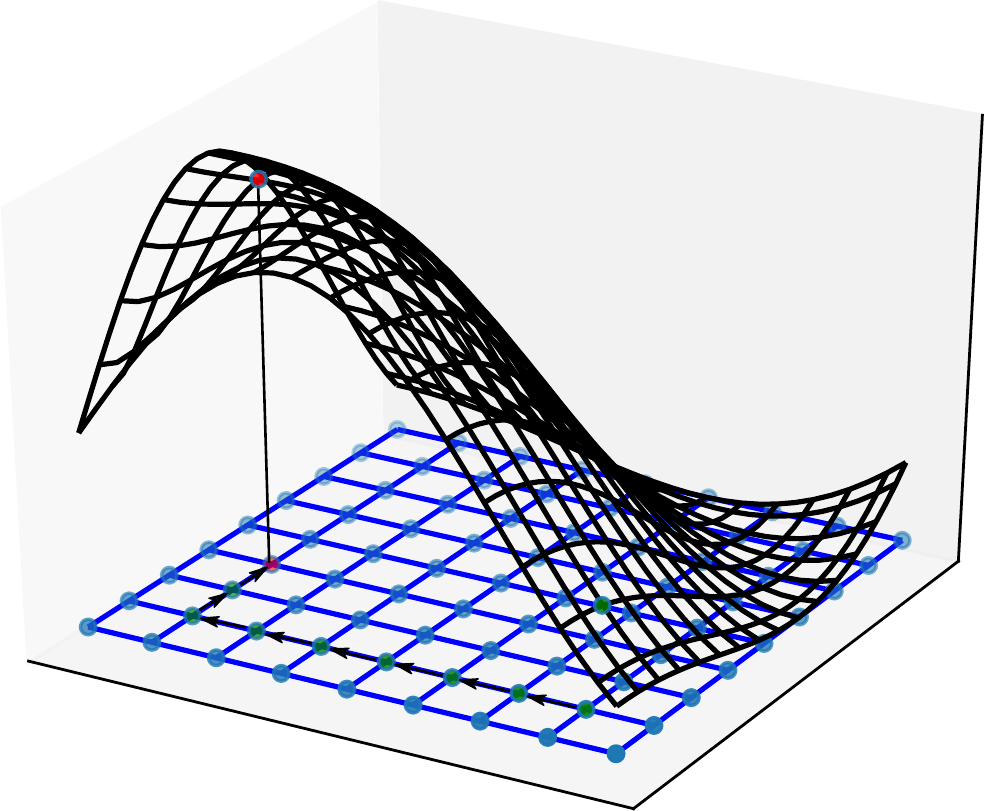} 
\caption{ Optimization of a black box function over a discrete, non-differentiable input space. The path from an initial input point highlighted in green is highlighted with arrows ending at the red input with optimal value.}
\label{fig:blackbox-discrete}
\end{figure}

\subsection{Contributions}

In this work, we develop a simple and efficient Bayesian optimization method for discrete combinatorial spaces by treating discrete variables as random variables and applying continuous relaxations. By employing the reparameterization trick combined with Thompson sampling and variational methods, we are able to apply gradient based optimization to maximize the acquisition function when finding new exploration data. Our method performs better than the current state-of-the-art approaches that we compared to and is highly scalable, since we use a Bayesian neural network surrogate model and gradient-based optimization in the algorithmic inner loop.

\section{Related Work}

There has been substantial work done in tackling discrete optimization, here we detail a few different approaches in Bayesian optimization and 
related methods. 

\noindent\textbf{Variational Optimization} 

There has been interesting relevant work treating discrete and non-differentiable optimization spaces by considering smoothed variational approximations of those spaces like in \cite{staines2012variational, Wierstra_2008}. 

\noindent\textbf{Treating discrete spaces as continuous}

A basic BO approach to combinatorial inputs is to represent all variables using one-hot
encoding, treating all integer-valued variables as values on a real line 
so that the acquisition function considers the closest integer for the chosen real value. 
This approach seems better suited for ordinal variables and while it has been used in low dimensional situations
\cite{Garrido_Merch_n_2020}, it has not proved useful for high dimensional ones. 

\noindent\textbf{Random search and evolutionary algorithms}

Methods like local search and evolutionary algorithms such as particle search are able to handle
black-box functions in discrete spaces. However,
these procedures have a variety of problems and are not designed to be sample efficient
and hence often prohibitively expensive. 
Moreover, local search algorithms do not necessarily converge to a global optimum. 
Other popular techniques such as mathematical programming, e.g.,
linear, convex, and mixed-integer programming, cannot be applied to black-box functions.

\noindent\textbf{Bayesian Optimization with sparse Bayesian linear regression}

In 2018, BOCS \cite{baptista2018bayesian} was proposed using sparse Bayesian linear regression instead of GPs. 
The acquisition function was optimized by a semi-definite
programming or simulated annealing that allowed to speed up the procedure of picking new points
for next evaluations. However, BOCS has certain limitations which restrict it to
problems with low order interactions between variables.

\noindent\textbf{BO with graph kernels}

In 2019, COMBO \cite{oh2019combinatorial} was proposed, it quantifies “smoothness” of functions on 
combinatorial search spaces by utilizing a combinatorial graph with a ARD diffusion kernel to 
model high-order interactions between variables which thus lead to better performance

\noindent\textbf{BO on attributed graphs}

Deep Graph Bayesian Optimization on attributed graphs was by proposed by \cite{Cui_2019}. They use deep graph neural network to model black-box functions on graph avoiding the cubic complexity of GPs that scales linearly with the number of observations. They test their method on molecular discovery and urban road network design. 

\noindent\textbf{Maximizing acquisition functions} 

\cite{wilson2018maximizing} show that acquisition functions estimated
via Monte-Carlo are amenable to gradient-based optimization. They also identify a common family of acquisition functions, including EI and UCB, whose properties lend them to greedy maximization approaches.

\noindent\textbf{Other ML approaches to Combinatorial Optimization} 

Work has also been done to tackle combinatorial optimization problems using neural networks and reinforcement learning, in \cite{bello2016neural}, focusing on the traveling salesman problem, they train a recurrent neural network that, given a set of city coordinates, predicts a distribution over different city permutations using negative tour length as the reward signal.

\begin{figure}[t]
\centering
\includegraphics[width=0.99\columnwidth]{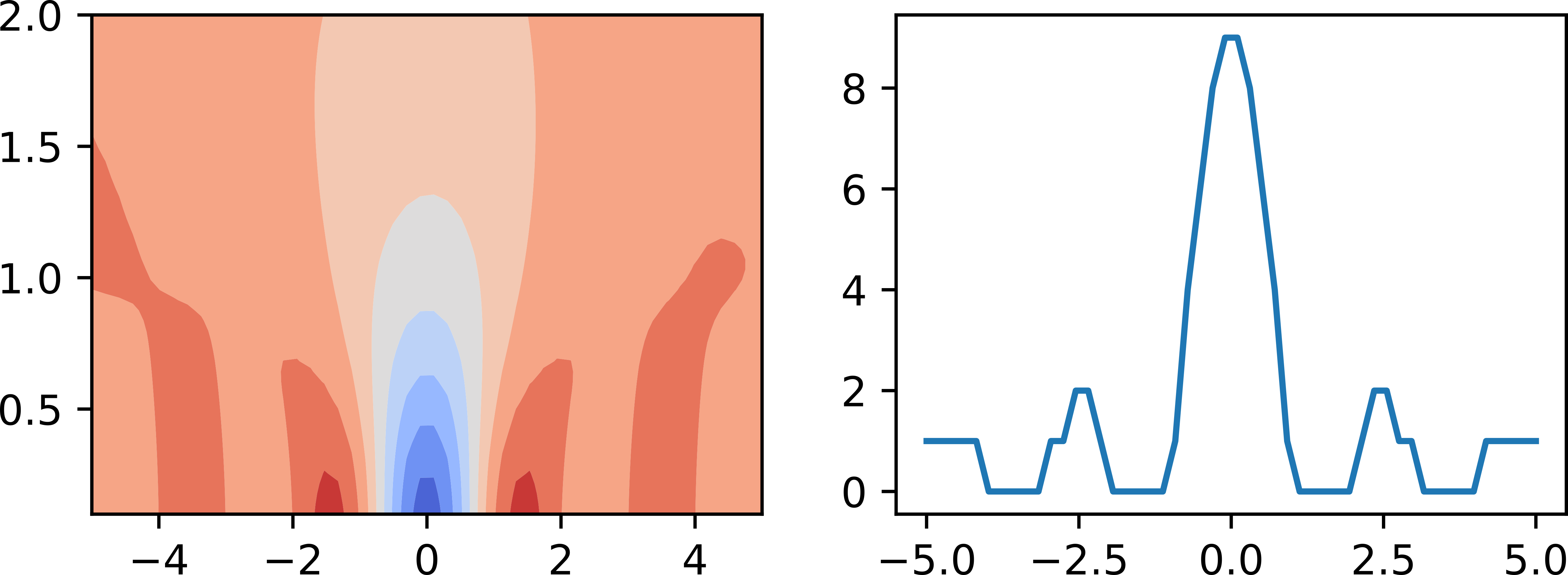} 

a) \hspace{4cm} b)

\caption{Variational approximation of a discrete function b) using a Gaussian $q(\B x |\bm \alpha)$ making it amenable to gradient based optimization.
a) is the contour map of the objective $\E_{q(\B x |\bm \alpha) }[f(\B x)]$.}
\label{var_opt}
\end{figure}

\section{Background}

We give a brief overview of variational and bayesian optimization before introducing our method

\subsection{Variational optimization}

Variational optimization \cite{staines2012variational, louppe2017adversarial} is a general optimization technique used to form a differentiable
bound on the optima of a non-differentiable function. Given our acquisition function $\elbo_{ \texttt{acq}} (\B x)$ to optimize, we can form a bound:
\begin{align*}
    \min _{\B x} f (\B x) \leq \E _{\bm x\sim q(\bm x |\bm \alpha)} 
[f(\B x)]
\end{align*}
where $q (\bm x |\bm \alpha)$ is a proposed distribution with parameters $\bm \alpha$ over input values $\B x$. 
That is, the minimum of the set of acquisition function values is always less than or equal to
any of the average function value. Provided that $q (\bm x |\bm \alpha)$ is flexible enough, the parameters $\bm \alpha $
can be updated to place its mass arbitrarily tight around the optimum. 

\begin{figure*}[t]
\centering
\includegraphics[width=0.99\textwidth]{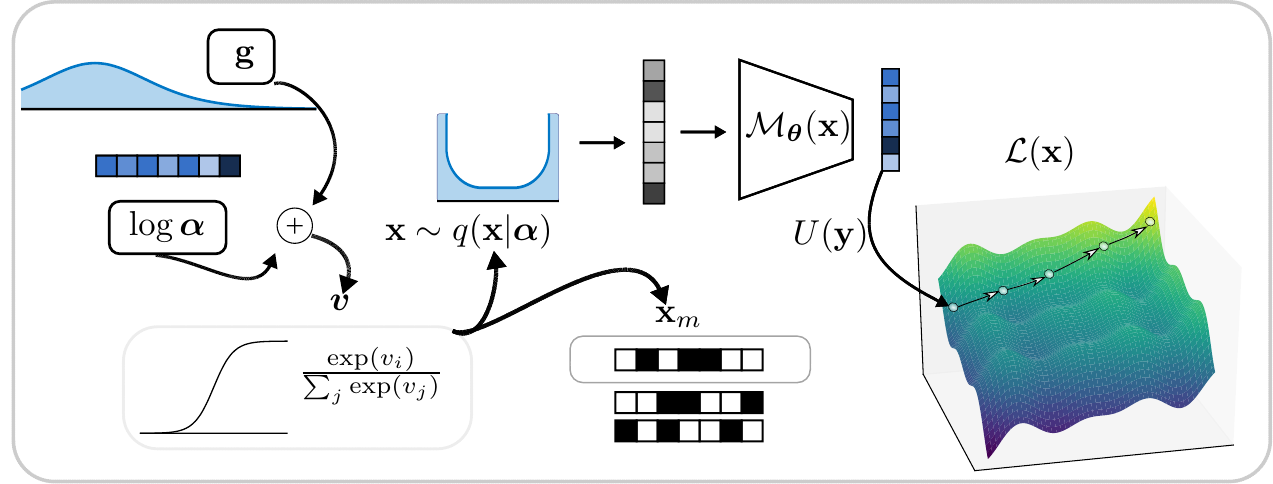}
\caption{The main steps of the Bayesian variational optimization (BVO) algorithm proposed in this work. This depicts the relaxation of the discrete search space and the direct optimization of the acquisition function by using a variational proposal $q(\B x | \bm \alpha)$ to find the optimal point in the input space $\B x_m$. }
\label{fig:bvo_graph}
\end{figure*}

\subsection{Bayesian optimization}

Bayesian optimization relies on both a surrogate model
$\mathcal{M}_{\bm\theta}$ and an acquisition function 
$\elbo_{ \texttt{acq}} (\B x)$ to define 
a strategy for efficiently maximizing a black-box function $f$. At each “outer-loop”
iteration, this strategy is used to choose a set  $\B x$ whose evaluation 
improves the search procedure. 

\noindent\textbf{Surrogate model}

The surrogate model $\mathcal{M}_{\bm \theta}$ provides a probabilistic interpretation of the underlying function $ f_k \sim p( f|\D)$. 
Gaussian processes are a natural surrogate model but as mentioned previously, their cost scales as the cube of the number of data. 


\noindent\textbf{Acquisition functions} are expectations over predictive distributions (of new outcomes) $p (\B y|\B x, \D) $
revealed when evaluating a black box function $f(\B x)$. 
This formulation naturally occurs in a Bayesian framework whereby $\B x $ is determined by accounting
for the utility provided by possible outcomes $ \B y \sim p(\B y|\B x, \D) $. Denoting the chosen utility function
as $U (\B y )$ we can represent the acquisition functions as 
\begin{equation}
\begin{split}
\elbo_{ \texttt{acq}} (\B x)
&= \E _{p(\B y | \B x,  \D )} [U ( \B y )] = \int p(\B y | \B x,  \D ) U ( \B y ) d\B y   
\end{split}
\end{equation}


\section{Problem Definition}

Given a black box function $f$ that is defined over a discrete structured domain $\mathcal{X}$ of feasible points, our goal
is to find a global optimizer $$ \B x^* = \argmin{\B x \in \mathcal{X} } f(\B x). $$

We focus on two main scenarios with search spaces consisting of 
\begin{itemize}
    \item binary variables with $\B x \in  \mathcal{X} = \{0, 1\}^d $, where $\B x_i$ equals one if a certain element $i$ is present or zero otherwise, and
    \item categorical variables  $\B x \in  \mathcal{X}  =\{0,1,2,\dots ,k\}^d$ where $\B x_{i}$ is a selected category from $1-k$ or zero if not present.
\end{itemize}
For example, we can generally associate a binary variable with an edge in a graph-like structure.

\section{Bayesian variational optimization}

The proposed Bayesian variational optimization (BVO) algorithm is a simple combination of three main parts: 
\begin{itemize}
    \item A Bayesian neural network surrogate model with Thompson sampling to approximate the predictive distribution. This allows us to form an unbiased estimate of the acquisition function.
    \item Variational optimization \cite{staines2012variational} of the acquisition function.
    \item A continuous relaxation of the search space for gradient-based 
    optimization of the acquisition function through a categorical reparameterization \cite{maddison2016concrete}.
\end{itemize}
The BVO algorithm are detailed in the next two sections and summarized graphically in 
figure \ref{fig:bvo_graph} as well as in algorithm \ref{alg:algo1} . 

\subsection{Bayesian Neural Network surrogate}
Flexible function approximation with reasonable uncertainty quantification can be done with 
Bayesian neural networks \cite{wunn}. Recent work \cite{hernndezlobato2017parallel}, in bayesian optimization has made successful use of BNNs as a surrogate model and we further apply these models in this work.

In BNNs, we suppose that observing $\B x$ provides independent, conditional, normal observations with mean $\mathcal{M}_{\bm \theta}(\B x)$ and finite variance
$\sigma^2$, such that we have total likelihood
$$p ( \D | \bm \theta ) = \prod_{\B x \in \D} 
\N(\mathcal{M}_{\bm \theta}(\B x),\sigma^2).$$ 
We approximate the true posterior $ p(\bm \theta | \D ) $ with a
diagonal multivariate Gaussian  
$$ q(\bm \theta) =\prod_i q_{\bphi} (\bm \theta _i) =\N (\bm \theta | \bm \mu , \bm \sigma ^2).$$
By making use of the reparameterization trick \cite{vae} $\bm \theta = \bm \mu + \bm \sigma \odot \bm \epsilon$, 
we can stochastically maximize a lower bound on the marginal log-likelihood
\begin{align}  
   \log p(\D ) & \geq \E_{q(\bm \theta)} \BB{ \log \frac{p( \bm \theta , \D)}{q(\bm \theta)}}.
\end{align} 

We use Thompson sampling to approximate the acquisition function 
with a single sample from the approximate posterior 
$\bm \theta \sim q(\bm \theta)$ so that 
$$ p(\B y | \B x , \D ) = \int  p(\B y | \B x, \bm \theta) p(\bm \theta |\D ) d\bm \theta \approx p(\B y | \B x, \bm \theta).$$

\begin{figure}[t]
\centering
\includegraphics[width=0.99\columnwidth]{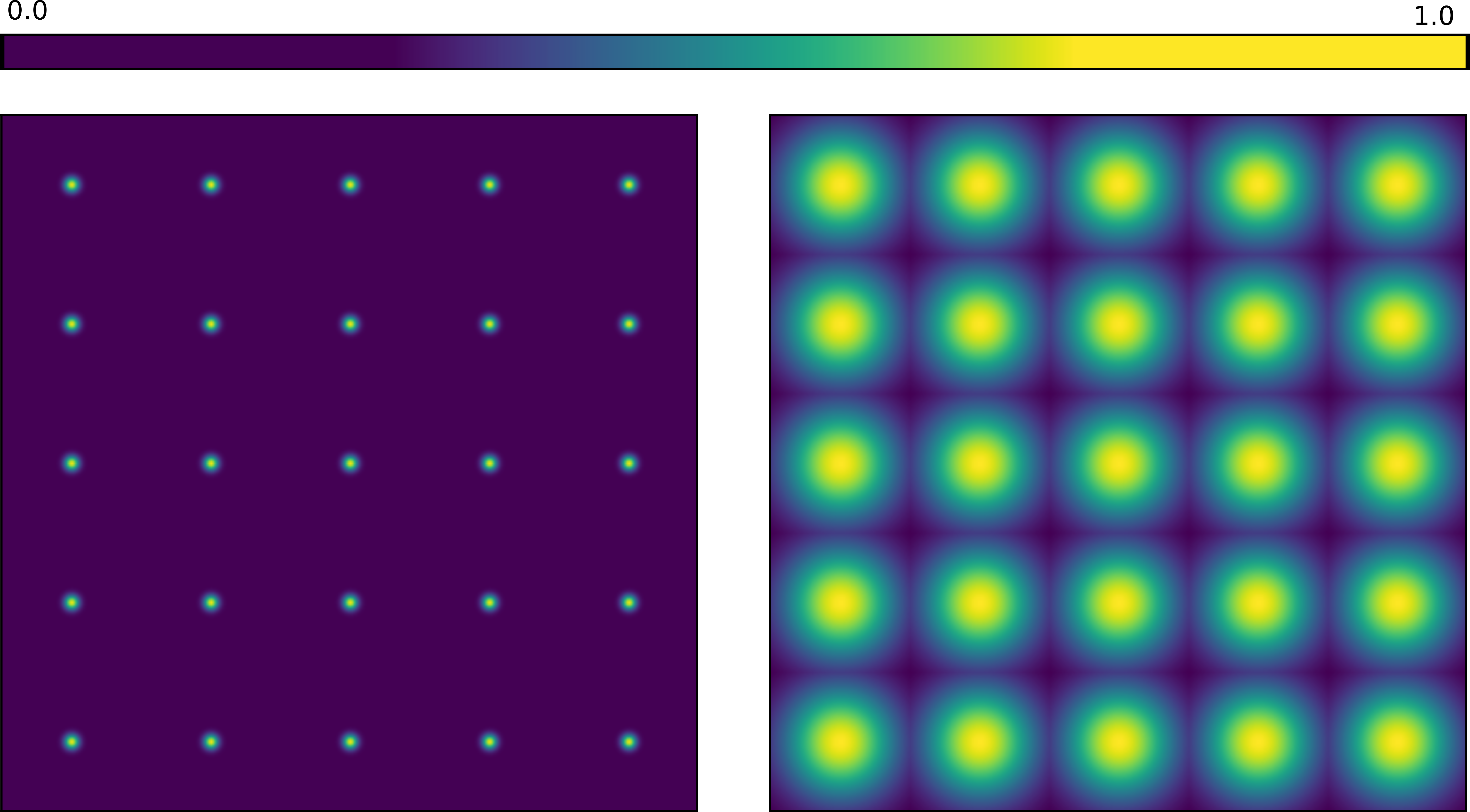} 

a) \hspace{4cm} b)

\caption{Relaxing a discrete space from figure \ref{fig:blackbox-discrete} with Gaussian noise to directly optimize some function: a) is the non-differentiable discrete space, and b) is the differentiable input space of an the function using Gaussians for $q(\B x |\bm \alpha)$.}
\label{fig:relaxation}
\end{figure}

\subsection{Variational optimization of the acquisition function}

Since the input to the acquisition function $ \elbo_{ \texttt{acq}}$ is discrete it is non-differentiable and we cannot optimize it with gradients. However, we can use variational optimization \cite{staines2012variational, louppe2017adversarial} to form a differentiable bound on 
$ \min _{\B x} \elbo_{ \texttt{acq}} (\B x) \leq \E _{\bm x\sim q(\bm x |\bm \alpha)} [\elbo_{ \texttt{acq}} (\B x)] $ and optimize that in place 
\begin{align*}
    \min _{\B x} \elbo_{ \texttt{acq}} (\B x) \leq \iint q(\bm x |\bm \alpha) (\B x)  p(\B y | \B x,  \theta  ) U ( \B y ) d\B y d\B x
\end{align*} 
using a single sample Monte-Carlo estimator of $ p(\B y | \B x , \D )$ and placing a proposal distribution $q (\bm x |\bm \alpha)$ on the input space $\B x$ with parameters $\bm \alpha$. Now we can optimize this input space by proxy through optimization of the proposal $q (\B x |\bm \alpha)$. 
Now we must consider what is an appropriate proposal distribution $q (\bm x |\bm \alpha)$ for example in the discrete space of figure \ref{fig:blackbox-discrete} 
we could use a Gaussian distribution as depicted in figure \ref{fig:relaxation}. Most search spaces we consider will consist of binary or categorical variables, hence it makes sense to model $q (\bm x |\bm \alpha)$ as a product of Bernoulli's or Multinomial distributions.

\subsection{Continuous Relaxation of the input space through Categorical Reparameterization}

To optimize the acquisition function directly with Monte-Carlo gradients, we are required to choose the proposal $q(\bm x |\bm \alpha)$ 
so that we can form
a differentiable reparameterization such that $\bm x = \eta (\bm \alpha, \bm \epsilon)$ \cite{kingma2013auto} and hence the
path-wise gradients from $\elbo_{ \texttt{acq}} (\B x(\bm \alpha))$ to $\alpha$ can be computed as
\begin{equation}
\begin{split}
   \frac{\partial}{\partial \bm \alpha } \E _{ \bm x\sim q(\bm x |\bm \alpha) } [\elbo_{ \texttt{acq}} (\B x)]
   &= \frac{\partial}{\partial \bm \alpha } \E _{\bm \epsilon \sim p(\bm \epsilon)} [\elbo_{ \texttt{acq}}(\eta(\bm \alpha ,\bm  \epsilon))] \\
   &= \E _{\bm \epsilon \sim p(\bm \epsilon)} \left [  \frac{\partial \elbo_{ \texttt{acq}}}{\partial \eta } \frac{\partial \eta}{\partial \bm \alpha } \right ].
\end{split}
\end{equation}
where $p(\bm \epsilon)$ is a base noise distribution. 
Motivated by this requirement, we use the concrete distribution \cite{maddison2016concrete} as our proposal distribution, 
which allows us to form a continuous approximation of categorical random variables 
by using the the Gumbel-Softmax trick. 
We can sample from the concrete distribution via  
\begin{align*}
    &\B g = -\log (-\log \B u ), \  \B u  \in [0, 1]^D , \\ 
    &\B x = \text{Softmax}((\log \bm \alpha +\B g )/\lambda).
\end{align*}
$\bm g $ is drawn from the Gumbel distribution with $\bm u $ drawn from
uniform distribution between $(0, 1)$. For the binary case, 
we can sample from the binary concrete distribution via  
\begin{align*}
    g &= \log u - \log (1-u ),  u  \sim  \mathcal{U}(0, 1) \\
    x &= \text{Sigmoid}((\log  \alpha +g )/\lambda).
\end{align*}
$ g $ is drawn from the logistic distribution with $u$ drawn from
uniform distribution between $(0, 1)$. In both cases, $\lambda$ is the temperature. 
The advantage for using the Gumbel-Softmax instead of Softmax for
discrete/categorical variables is that, with certain temperature,
the sampled distribution can be closer to one-hot representation.

\newcommand{\LL}{\mathcal{L}_{\texttt{acq} }}
\newcommand{\G}{\mathcal{G}}
\newcommand{\dif}[1]{\frac{\partial}{\partial #1}}
\begin{algorithm}[t]
    \DontPrintSemicolon\SetAlgoNoLine
    Given target function $f$, model $\M$, acquisition $\LL$, relaxation $\G$,
    and initial data $\mathcal{D}$\;
    \For{$i \leftarrow 1$ \KwTo $T$}{
        Fit model $\M$ with $\D$\;
        Randomly initialize a set of variables $\alpha$\;
        \For{$j \leftarrow 1$ \KwTo $N$}{
            Sample relaxed discrete variables 
            $\B X \leftarrow \G(\alpha)$\;
            Stochastic update 
            $\alpha \leftarrow \alpha +\gamma\dif{\alpha}\LL(\M(\B X))$\;
        }
        $\B X \leftarrow \G(\alpha)$\;
        Select max $\B x_m \leftarrow \argmax{\B x} \E[\LL(\M(\B X))]$\;
        Evaluate $\B y \leftarrow f(\B x_m)$\;
        Update $\D \leftarrow \D \cup (\B x_m, \B y) $
    }
    \caption{Bayesian Variational Optimization}
    \label{alg:algo1}
\end{algorithm}

\newcommand{\LMCOL}[1]{$\lambda=#1$}
\begin{table*}[t]
\centering
  \begin{tabular}{lcccccc}
    \toprule
    & \multicolumn{3}{c}{Ising} & \multicolumn{3}{c}{Contamination} \\
    \cmidrule(r){2-4}
    \cmidrule(r){5-7}
    Method & \LMCOL{0} & \LMCOL{10^{-4}} & \LMCOL{10^{-2}} & 
    \LMCOL{0} & \LMCOL{10^{-4}} &  \LMCOL{10^{-2}}  \\
    \midrule
    RS & 
    \PM{0.761}{0.643} & \PM{0.921}{0.755} & \PM{0.997}{0.689} & 
    \PM{21.92}{0.18} & \PM{21.87}{0.21} & \PM{22.03}{0.16} \\
    TPE* & 
    \PM{0.404}{0.109} & \PM{0.444}{0.095} & \PM{0.609}{0.107} & 
    \PM{21.64}{0.04} & \PM{21.69}{0.04} & \PM{21.84}{0.03} \\
    SA* & 
    \PM{0.095}{0.033} & \PM{0.117}{0.035} & \PM{0.334}{0.064} & 
    \PM{21.47}{0.04} & \PM{21.49}{0.04} & \PM{21.61}{0.03} \\
    BOCS-SDP* & 
    \PM{0.105}{0.031} & \PM{0.059}{0.013} & \PM{0.300}{0.039} & 
    \PM{21.37}{0.03} & \PM{21.38}{0.03} & \PM{21.52}{0.03} \\
    COMBO* & 
    \PM{0.103}{0.035} & \PM{0.081}{0.028} & \PM{0.317}{0.042} & 
    \BPM{21.28}{0.03} & \BPM{21.28}{0.03} & \BPM{21.44}{0.03} \\
    \midrule
    BVO & 
    \BPM{0.040}{0.059} & \BPM{0.050}{0.090} & \BPM{0.224}{0.057} & 
    \PM{21.33}{0.14} & \PM{21.37}{0.15} & \PM{21.49}{0.16} \\
    \bottomrule
  \end{tabular}
  \caption{Binary variable optimization over 25 runs.
  Baseline values * are from COMBO \cite{oh2019combinatorial}.}
  \label{tb:comparison}
\end{table*}

\section{Experiments}

To demonstrate the BVO algorithm, we preform several experiments to validate our approach in a variety of discrete spaces of variable size and structure, including:
\begin{enumerate}
    \item binary variables: 
    Ising sparsification and contamination control,
    \item categorical variables: 
    Pest control and molecular optimization with SELFIES,
    \item computational complexity on variable dimensions and
    data size. 
\end{enumerate}

\noindent\textbf{Optimization setup}

Due to the model-dependent nature of the
BVO method, we carried out a hyperparameter search for the
different objectives. This includes trying different 
BNN models, such as
various activation functions (Tanh, ReLU),
number of layers (2-6), 
and layer sizes (50-200). We also explored different 
acquisition functions (EI, SR, PI listed in \cite{wilson2018maximizing}), 
CONCRETE relaxation
temperatures (0.1-1.0), optimization batch size (16-512),
and the scaling factors ($10^1-10^7$) 
on the loss function.
The training of the models and optimization
were computed on Nvidia V100SXM2 GPUs from cloud servers with 8 core CPUs and
32 GB of RAM.

\noindent\textbf{Experimental baselines}

In our experimental demonstration of BVO, 
we compare with the following baseline algorithm on discrete optimization problems,
\begin{itemize}
    \item COMBO \cite{oh2019combinatorial},
    \item random search (RS), 
    \item simulated annealing (SA) \cite{spears1993simulated}, 
    \item Bayesian optimization of combinatorial structure (BOCS) \cite{baptista2018bayesian},
    \item and tree of Parzen estimators (TPE) \cite{bergstra2013making}.
\end{itemize}

\subsection{Binary variables}

\begin{figure}[ht]
    \centering
    \includegraphics[width=0.99\columnwidth]{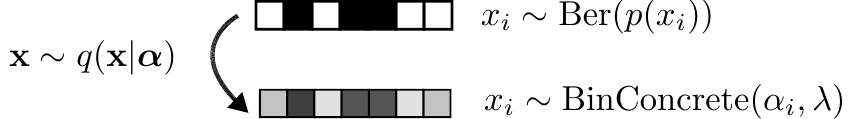}
    \caption{Treatment of binary spaces.}
    \label{fig:binary_spaces}
\end{figure}

Binary variable optimization is a ubiquitous problem in computer science. 
Therefore, the benchmarking of BVO with respect to this problem 
gives us a handle to compare to several methods from distinct areas of computer science.
For the binary variable optimization tasks, 
we benchmark the BVO method for 
Ising sparsification and contamination control tasks.
Figure \ref{fig:binary_spaces} shows our treatment of binary variables, 
and how we construct $q(\B x |\bm \alpha)$ as a product of Bernoulli distributions
that are relaxed for gradient based optimization. 

\noindent\textbf{The Ising sparsification problem}
entails the deletion of
interactions in a zero-field Ising model while 
maintaining a similar probability distribution to that of the
original model. The zero-field Ising model can be expressed as 
$$p(\bm{z})=\frac{1}{Z_p}\exp(\bm{z}^\top J^p\bm{z}),$$ 
where $\bm{z}\in \{-1, 1\}^n$, 
the interaction matrix $J^p \in \R^{n\times n}$, 
and $Z_p$ is the normalization factor. 
For this problem, our goal is to find an approximate
model $q(\bm{z})$ with an interaction matrix that are 
partially dropped out $J^q_{ij}=x_{ij}J^p_{ij}$, where
$x_{ij}\in \{0, 1\}$. For the model $q(\bm{z})$ to approximate
to $p(\bm{z})$, the objective function for optimization is
$$\mathcal{L}(\bm{x})=D_{KL}(p(z)||q(z))+\lambda ||x||_1,$$
where the first term is a Kullback-Leibler divergence for
distribution similarity and the second term is for maximizing dropouts.
In our benchmark, we use the same objective function in the
COMBO code base, which is a $4\times4$ Ising grid. Since the
interactions are only the nearest neighbors, there consiste only
24 possible interaction dropouts. All the interactions
$J^p_{ij}$ are sampled randomly from interval $[0.05, 0.5]$. 
The optimization includes 20 randomly initialized points followed by 
150 points of iterative selection for evaluation.

\begin{figure*}[t]
    \centering
    \includegraphics[width=0.95\linewidth]{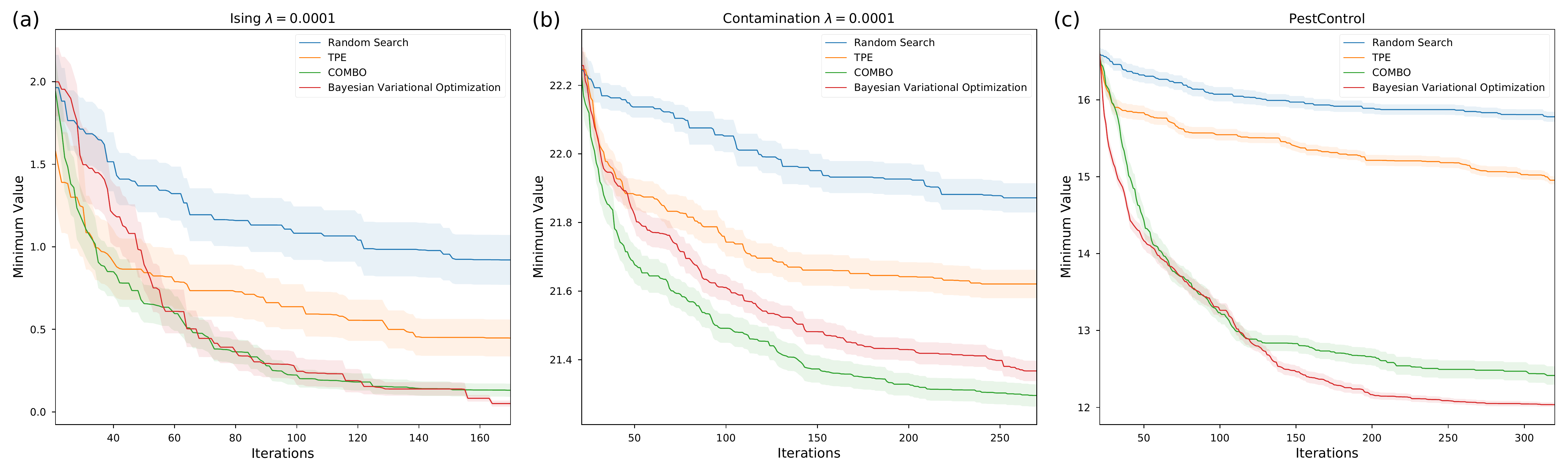}
    \caption{Bayesian variational optimization over 25 runs.
    The standard deviation in the plots are reduced by a factor of 5
    for viewing purposes.
    The comparisons are re-evaluated with 
    COMBO \cite{oh2019combinatorial}
    baseline codes and could be different to the 
    results in table \ref{tb:comparison}. The optimization values
    by iterations are plotted for (a) Ising sparsification problem 
    ($\lambda=10^{-4}$), 
    (b) Contamination control ($\lambda=10^{-4}$), and (c) Pest control.}
    \label{fig:bvotrend}
\end{figure*}

\noindent\textbf{Contamination control.}
The contamination control problem is a binary optimization problem for 
minimizing contamination in a simulated food supply chain.
At each stage $i$, $z_i\in[0,1]$ represents the portion 
of contaminated food, 
$$z_i = \alpha_i (1-x_i)(1-z_{i-1})+(1-\Gamma_ix_i)z_{i-1}.$$
where as the binary control variables $x_i\in\{0,1\}$ represents
whether to control and decontaminate the food. Contamination
at each stage may have different costs $c_i$. 
When no control $x_i=0$, the contaminated portion of the food in 
next stage will increase; where as $x_i=1$, certain portion of
the contaminated food will be decontaminated. The overall
target of the contamination control is to minimize
\begin{align} \label{eqn:conta_target}
    \mathcal{L} (\bm{x}) = \sum_{i=1}^d \left [  c_ix_i + 
    \sum_{k=1}^T  \frac{\rho}{T} 1_{\{z^k_i>u\}} \right ]  + 
    \lambda || x||_1
\end{align} 
The first term of the target is the cost of decontamination, and the 
last term is for $x$ regularization.
The second term is the penalty for every stages 
that have contaminated food over a certain threshold $u$, 
and it is calculated for the mean
of multiple starting food contamination portion $z^k_1$. In this
benchmark, we follow the COMBO baselines by using $T=100$,
$u=0.1$, with 100 food contamination stages and d=21 binary variables. The optimization
were 20 random initializing points with 250 optimizations.

\noindent\textbf{Results.}
The results for Bayesian variational optimization
on binary variables are shown in table 
\ref{tb:comparison}.
In the binary optimization problem, we can see
that BVO is similar or better in some cases
to state-of-the-art discrete optimization methods. 
However, in both Ising sparsification and contamination
control problems, BVO has a higher standard deviation
over the runs. This could be due to variational
sampling for optimization. 
The optimal values by optimization iterations are plotted in 
figure \ref{fig:bvotrend} (ab) for
Ising sparsification and Contamination control, which also includes
the baseline comparisons.

\subsection{Categorical variables}

\begin{figure}[ht]
\centering
\includegraphics[width=0.95\columnwidth]{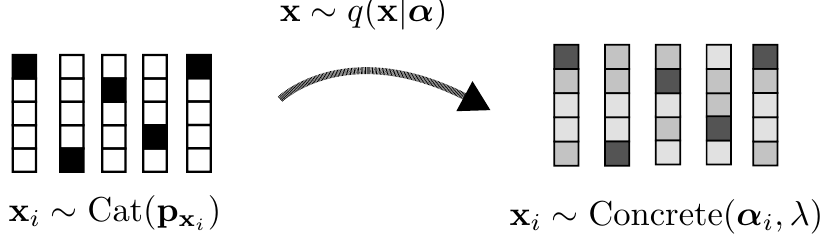}
\caption{Treatment of categorical spaces.}
\label{fig:cat_spaces}
\end{figure}

For the categorical variable (figure \ref{fig:cat_spaces}) optimization tasks, 
we benchmark our method to other baseline methods 
on the Pest control task. 
To demonstrate BVO in applications,
we aim to directly optimize molecules with
SELFIES representation.

\begin{figure}[t]
\centering
\includegraphics[width=0.95\columnwidth]{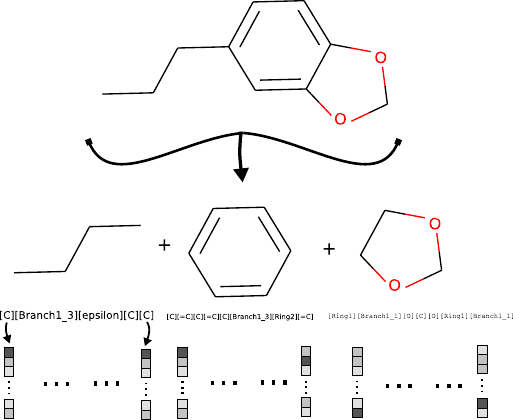}
\caption{Sequential representation of molecules using SELFIES, 
we convert a sequence of SELFIES tokens into a sequence of one-hot encodings 
modeling each token of the molecule as a relaxed categorical.}
\label{fig:selfies}
\end{figure}

\noindent\textbf{Pest control.}
The Pest control problem is similar to contamination control,
but with more control choices (4 different pesticide $l$).
The pest situation is also updated with a similar dynamics
to the contamination control problem
$$z_i = \alpha_i (1-x^l_i)(1-z_{i-1})+
(1-\Gamma^l_ix^l_i)z_{i-1}$$
but with different pesticide efficacy. However, there are also more
changes in dynamic control rates, spread rates,
and an reduce cost changes when a pesticide are used more often.
For this benchmark, there are 5 different categories
(no action + 4 pesticides) and 21 pest control stages. 
The optimization target is similar to 
equation \ref{eqn:conta_target}, without the regularization
term. For optimization, we use 20 random initial points
with 300 optimizations.

The results of pest control optimization are shown in table \ref{tb:pestcontrol}
along with the baseline comparisons.
Our method is on par to
the COMBO optimization results, while better than
the other methods. The minimum values by optimization
iterations are plotted in figure \ref{fig:bvotrend} (c) along
random search, TPE and COMBO, where we see that BVO performs
very well in terms of final optimal values and variances.

\begin{table}[ht]
\centering
  \begin{tabular}{lc}
    \toprule
    Method & Pest Control \\
    \midrule
    RS & \PM{15.779}{0.328} \\
    TPE* & \PM{14.261}{0.075} \\
    SA* & \PM{12.715}{0.091} \\
    COMBO* & \BPM{12.001}{0.003}\\
    \midrule
    BVO & \BPM{12.010}{0.027}\\
    \bottomrule
  \end{tabular}
  \caption{Pest control optimization over 25 runs.
  Baseline values * are from COMBO \cite{oh2019combinatorial}.}
  \label{tb:pestcontrol}
\end{table}

\begin{table}[ht]
\centering
  \begin{tabular}{lll}
    \toprule
    Method  & Highest score \\
    \midrule
    BVO with SELFIES  & \textbf{4.94}\\
    \midrule
    Random with SELFIES  & 3.47\\
    GVAE & 2.94 \\
    CVAE & 1.98 \\
    \bottomrule
  \end{tabular}
  \caption{Penalized LogP optimization with SELFIES.}
  \label{tb:selfies2}
\end{table}

\noindent\textbf{Molecule optimization with SELFIES.}
Molecular graphs are often represented as SMILES in chemistry,
but generating correct SMILES representation of a valid
molecules are difficult, due to both physical meaning
and grammar constraints. This could lead to low generative
validity, or complicated reinforcement learning algorithms
for generating the molecules. SELFIES \cite{krenn2019selfies}
is an improved version of SMILES representation, which 
encodes the graph in Chomsky type-2 context-free grammar 
that generates 100\% valid molecules.
Figure \ref{fig:selfies} shows an example on how a molecule
is represented in SELFIES.

In this optimization, our search space is 200 SELFIES
tokens, each token with 25 categories. The tokens include
construction of rings, branches, atoms and its bonds.
For the molecular target, we optimize for the highest penalized 
LogP score (solubility).
The search space for molecule optimization in this case 
is on the 
order of $10^{279}$. This scale is very difficult for 
GP based optimization methods.

Due to the large search space and complicated
sequence dependencies in SELFIES, we optimize the molecule
by initializing 2048 SELFIES strings, and parallel evaluating 512
samples for each optimization for 1200 steps. The model
is built with 1D convolution layers and pooling layers for
long sequence learning. The optimization were repeated over
10 times. The maximum score achieved for BVO is shown
in table \ref{tb:selfies2}, with a maximum score of 
4.94. As a comparison, GVAE have a maximum score of 2.94 and 
CVAE with 1.98.
This advantage for BVO with SELFIES could be due to the fact that 
our molecular search space can access outside of the ZINC dataset.
Nevertheless, the BVO on SELFIES has a better optimization LogP score
compared to a random search using SELFIES.


\begin{figure}
    \centering
    \includegraphics[width=\linewidth]{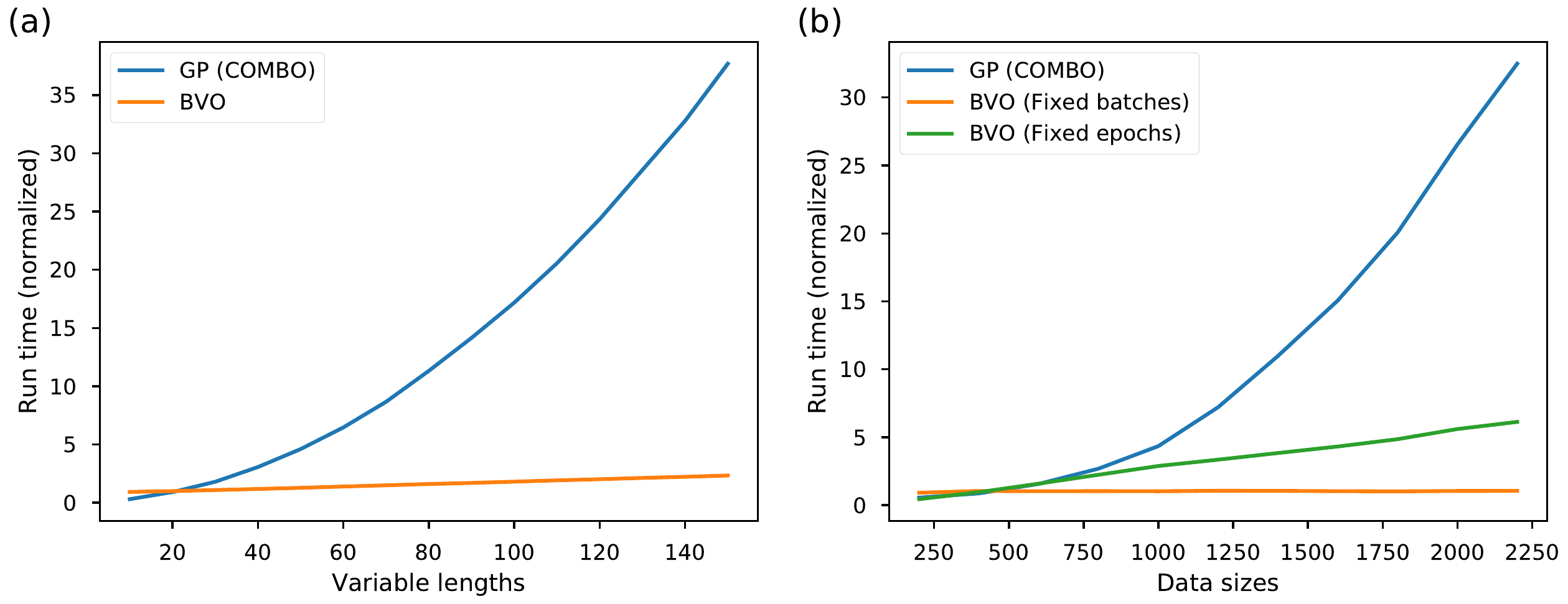}
    \caption{Complexity analysis of Bayesian variational
    optimization. 
    The run times are normalized to the first 3 data points. 
    (a) Computation time dependency on data variable lengths. 
    (b) Computation time dependency on total data size. 
    For the BVO, 
    we experimented on both fixed batches per prediction, and fixed number of epochs. }
    \label{fig:complexity}
\end{figure}

\subsection{Scalability of BVO}

Bayesian optimization often relies on Gaussian processes (GP)
for the surrugate model. However, GP has the limitation
that computation costs scale cubically to data size. 
This may lead to difficulty on categorical optimization tasks
when they have high variable dimensions
and requires large data size for training. 
In figure \ref{fig:complexity}, we measured the computation time
for BVO and GP (COMBO) on the pest control experiment. 
For comparison, the mean of first
3 points of each curve are normalized to 1. 
The computation scaling regarding to variable dimension
for BVO is linear scaling (order = 1.09), in contrast to GP 
which is quadratically $O(n^2)$ (order = 1.93). 
Regarding with data size (optimization iterations), 
GP scales between $O(n^2)-O(n^3)$ (order = 2.49), while
BVO can scale by constant (fixed number of batches per epoch) or by linear 
(fixed epochs of data size).

The complexity analysis here shows that BVO is more practical
then GP for optimizing large problems, which we demonstrated on SELFIES optimization.

\section{Conclusion}

In this paper, we developed a simple and efficient Bayesian variational optimization method for discrete combinatorial spaces. By employing reparameterization and Thompson sampling, we are able to apply gradient based optimization to maximize the acquisition function when finding new exploration data. While our method performs better than the current state-of-the-art approaches, the algorithm also allows high dimension and large data size optimization without suffering computation limitations. 
As a demonstration, we optimized the molecules with BVO on SELFIES representation, which consist of large search spaces.

\section{Acknowledgement}
Machine learning models are trained with the GPUs  on B\'{e}luga, Calcul Qu\'{e}bec. The training was enabled in part by support provided by Calcul Qu\'{e}bec (https://www.calculquebec.ca/) and Compute Canada (www.computecanada.ca).

\bibliography{reference}

\begin{thebibliography}{20}
\providecommand{\natexlab}[1]{#1}
\providecommand{\url}[1]{\texttt{#1}}
\providecommand{\urlprefix}{URL }
\expandafter\ifx\csname urlstyle\endcsname\relax
  \providecommand{\doi}[1]{doi:\discretionary{}{}{}#1}\else
  \providecommand{\doi}{doi:\discretionary{}{}{}\begingroup
  \urlstyle{rm}\Url}\fi

\bibitem[{Baptista and Poloczek(2018)}]{baptista2018bayesian}
Baptista, R.; and Poloczek, M. 2018.
\newblock Bayesian optimization of combinatorial structures.
\newblock \emph{arXiv preprint arXiv:1806.08838} .

\bibitem[{Bello et~al.(2016)Bello, Pham, Le, Norouzi, and
  Bengio}]{bello2016neural}
Bello, I.; Pham, H.; Le, Q.~V.; Norouzi, M.; and Bengio, S. 2016.
\newblock Neural Combinatorial Optimization with Reinforcement Learning.

\bibitem[{Bergstra, Yamins, and Cox(2013)}]{bergstra2013making}
Bergstra, J.; Yamins, D.; and Cox, D.~D. 2013.
\newblock Making a science of model search: Hyperparameter optimization in
  hundreds of dimensions for vision architectures .

\bibitem[{Blundell et~al.(2015)Blundell, Cornebise, Kavukcuoglu, and
  Wierstra}]{wunn}
Blundell, C.; Cornebise, J.; Kavukcuoglu, K.; and Wierstra, D. 2015.
\newblock Weight uncertainty in neural networks.
\newblock \emph{International Conference on Machine Learning} .

\bibitem[{Cui, Yang, and Hu(2019)}]{Cui_2019}
Cui, J.; Yang, B.; and Hu, X. 2019.
\newblock Deep Bayesian Optimization on Attributed Graphs.
\newblock \emph{Proceedings of the AAAI Conference on Artificial Intelligence}
  33: 1377–1384.
\newblock ISSN 2159-5399.
\newblock \doi{10.1609/aaai.v33i01.33011377}.
\newblock \urlprefix\url{http://dx.doi.org/10.1609/aaai.v33i01.33011377}.

\bibitem[{Garrido-Merchán and Hernández-Lobato(2020)}]{Garrido_Merch_n_2020}
Garrido-Merchán, E.~C.; and Hernández-Lobato, D. 2020.
\newblock Dealing with categorical and integer-valued variables in Bayesian
  Optimization with Gaussian processes.
\newblock \emph{Neurocomputing} 380: 20–35.
\newblock ISSN 0925-2312.
\newblock \doi{10.1016/j.neucom.2019.11.004}.
\newblock \urlprefix\url{http://dx.doi.org/10.1016/j.neucom.2019.11.004}.

\bibitem[{Gómez-Bombarelli et~al.(2018)Gómez-Bombarelli, Wei, Duvenaud,
  Hernández-Lobato, Sánchez-Lengeling, Sheberla, Aguilera-Iparraguirre,
  Hirzel, Adams, and Aspuru-Guzik}]{G_mez_Bombarelli_2018}
Gómez-Bombarelli, R.; Wei, J.~N.; Duvenaud, D.; Hernández-Lobato, J.~M.;
  Sánchez-Lengeling, B.; Sheberla, D.; Aguilera-Iparraguirre, J.; Hirzel,
  T.~D.; Adams, R.~P.; and Aspuru-Guzik, A. 2018.
\newblock Automatic Chemical Design Using a Data-Driven Continuous
  Representation of Molecules.
\newblock \emph{ACS Central Science} 4(2): 268–276.
\newblock ISSN 2374-7951.
\newblock \doi{10.1021/acscentsci.7b00572}.
\newblock \urlprefix\url{http://dx.doi.org/10.1021/acscentsci.7b00572}.

\bibitem[{Hernández-Lobato et~al.(2017)Hernández-Lobato, Requeima,
  Pyzer-Knapp, and Aspuru-Guzik}]{hernndezlobato2017parallel}
Hernández-Lobato, J.~M.; Requeima, J.; Pyzer-Knapp, E.~O.; and Aspuru-Guzik,
  A. 2017.
\newblock Parallel and Distributed Thompson Sampling for Large-scale
  Accelerated Exploration of Chemical Space.

\bibitem[{Jones, Schonlau, and Welch(1998)}]{Jones_1998}
Jones, D.~R.; Schonlau, M.; and Welch, W.~J. 1998.
\newblock \emph{Journal of Global Optimization} 13(4): 455–492.
\newblock ISSN 0925-5001.
\newblock \doi{10.1023/a:1008306431147}.
\newblock \urlprefix\url{http://dx.doi.org/10.1023/A:1008306431147}.

\bibitem[{Kingma and Welling(2013)}]{kingma2013auto}
Kingma, D.~P.; and Welling, M. 2013.
\newblock Auto-encoding variational bayes.
\newblock \emph{arXiv preprint arXiv:1312.6114} .

\bibitem[{{Kingma} and {Welling}(2014)}]{vae}
{Kingma}, D.~P.; and {Welling}, M. 2014.
\newblock {Auto-Encoding Variational Bayes}.
\newblock \emph{International Conference on Learning Representations} .

\bibitem[{Krenn et~al.(2019)Krenn, H{\"a}se, Nigam, Friederich, and
  Aspuru-Guzik}]{krenn2019selfies}
Krenn, M.; H{\"a}se, F.; Nigam, A.; Friederich, P.; and Aspuru-Guzik, A. 2019.
\newblock SELFIES: a robust representation of semantically constrained graphs
  with an example application in chemistry.
\newblock \emph{arXiv preprint arXiv:1905.13741} .

\bibitem[{Louppe, Hermans, and Cranmer(2017)}]{louppe2017adversarial}
Louppe, G.; Hermans, J.; and Cranmer, K. 2017.
\newblock Adversarial Variational Optimization of Non-Differentiable
  Simulators.

\bibitem[{Maddison, Mnih, and Teh(2016)}]{maddison2016concrete}
Maddison, C.~J.; Mnih, A.; and Teh, Y.~W. 2016.
\newblock The concrete distribution: A continuous relaxation of discrete random
  variables.
\newblock \emph{arXiv preprint arXiv:1611.00712} .

\bibitem[{Oh et~al.(2019)Oh, Tomczak, Gavves, and
  Welling}]{oh2019combinatorial}
Oh, C.; Tomczak, J.~M.; Gavves, E.; and Welling, M. 2019.
\newblock Combinatorial bayesian optimization using graph representations.
\newblock \emph{arXiv preprint arXiv:1902.00448} .

\bibitem[{Snoek, Larochelle, and Adams(2012)}]{snoek2012practical}
Snoek, J.; Larochelle, H.; and Adams, R.~P. 2012.
\newblock Practical Bayesian Optimization of Machine Learning Algorithms.

\bibitem[{Spears(1993)}]{spears1993simulated}
Spears, W.~M. 1993.
\newblock Simulated annealing for hard satisfiability problems.
\newblock \emph{Cliques, Coloring, and Satisfiability} 26: 533--558.

\bibitem[{Staines and Barber(2012)}]{staines2012variational}
Staines, J.; and Barber, D. 2012.
\newblock Variational Optimization.

\bibitem[{Wierstra et~al.(2008)Wierstra, Schaul, Peters, and
  Schmidhuber}]{Wierstra_2008}
Wierstra, D.; Schaul, T.; Peters, J.; and Schmidhuber, J. 2008.
\newblock Natural Evolution Strategies.
\newblock \emph{2008 IEEE Congress on Evolutionary Computation (IEEE World
  Congress on Computational Intelligence)} \doi{10.1109/cec.2008.4631255}.
\newblock \urlprefix\url{http://dx.doi.org/10.1109/CEC.2008.4631255}.

\bibitem[{Wilson, Hutter, and Deisenroth(2018)}]{wilson2018maximizing}
Wilson, J.; Hutter, F.; and Deisenroth, M. 2018.
\newblock Maximizing acquisition functions for Bayesian optimization.
\newblock In \emph{Advances in Neural Information Processing Systems},
  9884--9895.

\end{thebibliography}
\end{document}